\documentclass[journal]{IEEEtran}
\usepackage{microtype}
\usepackage{graphicx}
\usepackage{subfigure}
\usepackage{booktabs} % for professional tables
\usepackage{amsmath}
\usepackage{amsthm}	
\usepackage{amsfonts}

\usepackage[utf8]{inputenc} % allow utf-8 input
\usepackage[T1]{fontenc}    % use 8-bit T1 fonts
\usepackage{hyperref}       % hyperlinks
\usepackage{url}            % simple URL typesetting
\usepackage{booktabs}       % professional-quality tables
\usepackage{amsfonts}       % blackboard math symbols
\usepackage{nicefrac}       % compact symbols for 1/2, etc.
\usepackage{microtype}      % microtypography

\usepackage{amssymb}

\usepackage{fancyhdr}
\ifCLASSINFOpdf
\else
\fi
\begin{document}
\title{Understanding Global Loss Landscape of One-hidden-layer ReLU Networks\\ Part 2:  Experiments and Analysis}

\author{Bo Liu% <-this % stops a space
	\thanks{Bo Liu is with College of Computer Science, Faculty of Information Technology, Beijing University of Technology, Beijing, China. e-mail: liubo@bjut.edu.cn.}}

\maketitle

\begin{abstract}
The existence of local minima for one-hidden-layer ReLU networks has been investigated theoretically in \cite{globallosslandscape_part1}. Based on the theory, in this paper, we first analyze how big the probability of existing local minima is for 1D Gaussian data and how it varies in the whole weight space. We show that this probability is very low in most regions. We then design and implement a linear programming based approach to judge the existence of genuine local minima, and use it to predict whether bad local minima exist for the MNIST and CIFAR-10 datasets, and find that there are no bad differentiable local minima almost everywhere in weight space once some hidden neurons are activated by samples. These theoretical predictions are verified experimentally by showing that gradient descent is not trapped in the cells from which it starts. We also perform experiments to explore the count and size of differentiable cells in the weight space.
\end{abstract}


% \fi
\IEEEpeerreviewmaketitle

\section{Introduction}\label{section1}

\IEEEPARstart{I}{n} part 1 of this work \cite{globallosslandscape_part1}, we have studied the global loss landscape of one-hidden-layer ReLU networks from a theoretical persperctive. For one-hidden-layer ReLU networks, the space of weight vector is partitioned into a number of convex cells by input data samples. \cite{globallosslandscape_part1} proved that there are no bad local minima in the local landscapes inside cells, and gave the conditions for the existence of genuine local minima and their locations if they do exist. See section \ref{section2} for a detailed description. 

In this paper, based on the theory in \cite{globallosslandscape_part1}, to get an idea of how big the probability of existing bad local minima is at anyplace in the whole weight space, we first give an analytical investigation of this probability for 1D Gaussian data. We then describe how to implement an efficient approach to judge the existence of genuine local minima when they are in the form of hyperplanes. We conduct experiments on both synthetic and real datasets to predict the existence of bad local minima with our theory, as well as verify the correctness of these theoretical predications. Finally,  we will also carry out experiments to explore the count and size of differentiable cells in weight space to give a more complete picture of loss landscape.

More specifically, we have made the following contributions in this paper.

\begin{itemize}
	\item Design and implement intersection of half-spaces with linear programming to judge the existence of genuine local minima when they are in the form of hyperplanes.
	\item We show analytically that for 1D Gaussian data the probability of existing bad local minima is very low in most regions.  
	\item We use our theory to predict whether bad local minima exist for the MNIST and CIFAR-10 datasets, and find that there are no bad local minima in typical locations from far to near in weight space once some hidden neurons are active.		
	\item The theoretical predictions of whether bad local minima exist are verified experimentally by showing that gradient descent is not trapped in the cells it starts from. 
	\item We conduct experiments to explore the size of differentiable cells in weight space.
\end{itemize}

These theoretical and experimental results explain, from the viewpiont of loss landscape, why local search based methods such as gradient descent can optimize successfully one-hidden-layer ReLU networks of any size and any input if initialized at appropriate locations.

This paper is organized as follows. Section \ref{section2} gives a brief introduction to the theory of existence of genuine differentiable local minima for one-hidden-layer ReLU networks. In section \ref{section3}, we compute the probability of existing local minima for 1D Gaussian input, and demonstrate how this probability varies in weight space with experiments. In section \ref{section4}, we present implementation of intersection of half-spaces to judge the existence of genuine local minima when they are hyperplanes, and experiments on higher-dimensional Gaussian input. Section \ref{section5} presents experiments on MNIST and CIFAR-10 datasets, showing the consistency between theoretical predictions and experimental results. The count and size of convex cells are explored in section \ref{section6}. Section \ref{section7} is related work. Finally, we conclude this paper and point out future directions. 

\section{Preliminaries}\label{section2}
\subsection{Locations of Differentiable Local Minima}\label{section2.1}
Suppose there are $ K $ hidden neurons with ReLU non-linearality, \textit{d} input neurons and a single output neuron in a one-hidden-layer ReLU network. The input samples are $ (\mathbf{x}_i,\ y_i)\ (i\in [N]), $ where $ \mathbf{x}_i\in \mathbb{R}^d$ is data vector and $ y_i\in\pm1 $ is the label of  $ \mathbf{x}_i $, $[N]$ stands for $\left \{ 1,2,\cdots,N \right \}$. The loss of a one-hidden-layer ReLU network is
\begin{equation}\label{eq1}
L(z,\mathbf{{w}})=\frac{1}{N}\sum_{i=1}^{N}l(\sum_{j=1}^{K}z_j\cdot \begin{bmatrix}
\mathbf{w}_j\cdot \mathbf{x}_i
\end{bmatrix}_{+},y_i),
\end{equation}
where $z=\left\{z_k,\ k\in [K]\right\}$ are the weights between output neuron and hidden ones, $\mathbf{w}= \left\{\mathbf{w}_k, \ k\in [K]\right\}$ represent the weight vectors (augmented with bias) connecting hidden neurons and input, $ [y]_+=max(0,y) $ is the ReLU function and \textit{l} is the loss function.

For one-hidden-layer ReLU network model, $\mathbf{x}_i$ is a hyperplane in the weight space, and the samples $ \left \{ \mathbf{x}_i,\ i\in [N]  \right \}$ partition the weight space into a number of convex cells. Introducing variables $I_{ij}$ which equal 1 if $\mathbf{w}_j \cdot \mathbf{x}_i>0$ and 0 otherwise, and defining $\mathbf{R}_j=z_j\cdot\mathbf{w}_j$, the loss can be rewritten as 
\begin{equation}\label{A.2}
L(\mathbf{R})= \frac{1}{N}\sum_{j=1}^{N}{l(\sum_{j=1}^{K}{I_{ij}\mathbf{R}_j\cdot\mathbf{x}_i}},y_i) 
\end{equation}
where $ \mathbf{R}=\begin{pmatrix}
{\mathbf{R}_1}^T& \hdots &{\mathbf{R}_K}^T 
\end{pmatrix}^T$.

Given any $\left\{\mathbf{w}_j,\ j\in [K]\right\}$, each weight vector $\mathbf{w}_j$ will be located in a certain cell, and we call these cells in which $\left\{\mathbf{w}_j,\ j\in [K]\right\}$ reside as their \textbf{defining cells}. Inside defining cells, $\left \{ I_{ij},\ i\in [N], j\in [K] \right \} $ are constant, hence $ L(\mathbf{R}) $ is a differentiable function of $ \mathbf{R} $. It has been proved in \cite{globallosslandscape_part1} that there are no bad local minima in the local landscape of defining cells of any $\left\{\mathbf{w}_j,\ j\in [K]\right\}$ if $ l $ is convex. 

When loss function \textit{l} is squared loss, the location of local minimum for the cells specified by constant $\left \{ I_{ij},\ i\in [N], j\in [K] \right \} $ is given by the following solution
\begin{equation}\label{eq7}
\mathbf{R^{\ast }}=A^+\mathbf{y} + \left(I-A^+A\right)\mathbf{c},
\end{equation}
where $ \mathbf{c}\in \mathbb{R}^{Kd}$ is a arbitrary vector, $I$ is identity matrix.  $A^+$ is the Moore-Penrose inverse of $A$, and

\begin{equation}\label{eq6}
A=\left(\begin{matrix}I_{11}\mathbf{x}_1^T&\cdots& I_{1K}\mathbf{x}_1^T\\\vdots&\ddots&\vdots\\I_{N1}\mathbf{x}_N^T&\cdots& I_{NK}\mathbf{x}_N^T\\\end{matrix}\right),\mathbf{y}=\left(\begin{matrix}\begin{matrix}y_1\\y_2\\\end{matrix}\\\begin{matrix}\vdots\\y_N\\\end{matrix}\\\end{matrix}\right).
\end{equation}

The solution $\mathbf{R}^\ast$ can be characterized by the following cases:

1). $\mathbf{R}^\ast$ is unique: $\mathbf{R}^\ast = A^+\mathbf{y}$, corresponding to $A^+A=I$ and thus $(I-A^+A)\mathbf{c}=0$. This happens if and only if $rank (A)=Kd$. Therefore, $N\geq Kd$ is necessary in order to have a unique solution. 

2). $\mathbf{R}^\ast$ has infinite number of continuous solutions. In this case, $I-A^+A\neq0$, hence the arbitrary vector $\mathbf{c}$ plays a role. This happens only if $rank(A)\neq Kd$, corresponding to two possible situations. a). $N<Kd$. This is usually refered to as over-parameterization. b). $N\geq Kd$ but $rank(A)< Kd$.

In this case, \eqref{eq7} shows $\mathbf{R}^\ast$ is a affine transformation of $ \mathbf{c}\in \mathbb{R}^{Kd}$. Therefore, $\mathbf{R}_j^\ast$ can be the whole $\mathbb{R}^{d}$ space or a linear subspace of it (e.g. hyperplanes), depending on whether the rows in $\left(I-A^+A\right)$ corresponding to $\mathbf{R}_j^\ast$ is of full rank or not. 

The loss at local minima in \eqref{eq7} is
\begin{equation}\label{eq9}
L(\mathbf{R}^\ast)=\frac{1}{N}\left \| AA^+\mathbf{y}-\mathbf{y} \right \|_{2}^{2}
\end{equation}

\subsection{Criteria for Existence of Genuine Differentiable Local Minima}\label{section2.2}
Given any $\left\{\mathbf{w}_j,\ j\in [K]\right\}$ and corresponding defining cells, the locations of associated local minima are given in \eqref{eq7}. However, such locations may be outside the defining cells, and if so, there actually exist no local minima in these cells. We call those local minima that are still located in their defining cell as \textbf{genuine local minima}.

The conditions under which $\left\{\mathbf{R}^\ast_j,\ j\in [K]\right\}$ will be inside their defining cells are given as follows.

1). For the case $\mathbf{R}^\ast$ is unique, in order for $\mathbf{w}^\ast$ to be inside the defining cells, $\mathbf{w}^\ast$ and $\mathbf{w}$ should be on the same side of each sample . Giving $\left \{ I_{ij},\ i\in [N], j\in [K] \right \} $ that specify the defining cells, this can be expressed as
\begin{equation}\label{eq10}
\mathbf{w}_j^\ast\cdot\mathbf{x}_i \begin{cases} >0 \ \ \text{if } I_{ij}=1; \\ \le0 \ \ \text{if } I_{ij}=0; \end{cases} \ (i\in [N]; j\in [K]).
\end{equation}

Corresponding to different signs of $z_j^\ast$, the criteria for existing unique differentiable local minima can be expressed as: for each $\mathbf{R}_j^\ast\, (j\in [K])$,
\begin{equation}\label{eq11}
\mathbf{R}_j^\ast\cdot\mathbf{x}_i \begin{cases} >0 \ \ \text{if } I_{ij}=1; \\ \le0 \ \ \text{if } I_{ij}=0; \end{cases} \ (i\in [N])  
\end{equation}
\begin{equation}\label{eq12}
\ \ or \ \  \mathbf{R}_j^\ast\cdot\mathbf{x}_i \begin{cases} <0 \ \ \text{if } I_{ij}=1; \\ \ge0 \ \ \text{if } I_{ij}=0; \end{cases} \ (i\in [N])
\end{equation}

2). For the case $\mathbf{R}^\ast$ is continuous, we need to test whether the continuous differentiable local minima in \eqref{eq7} are in their defining cells. For example, substituting \eqref{eq7} into \eqref{eq11}, then for each $\mathbf{R}_j^\ast\ (j\in [K])$ the criteria become
\begin{gather}\label{eq13}
\mathbf{x}_i^T((A^+\mathbf{y})_j+(I-A^+A)_j\mathbf{c}) \begin{cases} >0 \ \ \text{if } I_{ij}=1; \\ \le0 \ \ \text{if } I_{ij}=0; \end{cases}  (i\in [N])
\end{gather}
where ${(A^+\mathbf{y})}_j$ is the rows of $A^+\mathbf{y}$ corresponding to $\mathbf{R}_j^\ast$, and so on. Each inequality of $\mathbf{c}$ in \eqref{eq13} defines a half-space in $\mathbb{R}^{Kd}$. Therefore, the criterion for existing genuine continuous differentiable local minima is transformed into identifying whether there exists non-null intersection of all these half-spaces. 

\section{Probability of Existing Genuine Local Minima for 1D Gaussian Data}\label{section3}
\subsection{Locations of Local Minima for 1D Gaussian Data}\label{section3.1}

In this section, in order to see how big the probability of existing genuine local minima is and how it varies in the weight space, we will compute the probability of existing genuine local minima analytically for 1D Gaussian input data. The core idea is that if no samples lie between the original weight vector $\mathbf{w}$ and the local minima $\mathbf{w}^\ast$, then \eqref{eq10} will hold and $\mathbf{w}^\ast$ will be inside the same cell with $\mathbf{w}$ and thus be a genuine local minimum. Therefore, the probability of existing genuine local minima is actually the probability of having no samples between $\mathbf{w}$ and $\mathbf{w}^\ast$. However, it is hard to get an analytical probability starting from the general solution of $\mathbf{w}^\ast$ in \eqref{eq7}. To get analytical solutions, we consider the simple case of 1D input, which is also equivalent to the case where all higher-dimensional weights are parallel. Higher-dimensional Gaussian input will be discussed in section \ref{section4}.

For the case of 1D input, denote the unit vector of $x$ coodinate as $\mathbf{i}$, a weight vector $\mathbf{w}_k$ is then represented by its normal $\mathbf{n}_k=\mathbf{i}$ or $\mathbf{n}_k=-\mathbf{i}$ and its $ x $ coordinate $h_k$. During optimization, we fix the normal of each weight vector and only tune its location. Furthermore, we fix the value of $z_k$ and set $z_k=1$ if $\mathbf{n}_k=\mathbf{i}$ and $z_k=-1$ if $\mathbf{n}_k=-\mathbf{i}$. As explained in \cite{globallosslandscape_part1}, the magnitude of $z_k$ does not matter for identifying existence of genuine local minima. Therefore, by the fact that $ \mathbf{w}_k\cdot \mathbf{x}_i$ is equal to the signed distance from $\mathbf{x}_i$ to $\mathbf{w}_k$, we have $\mathbf{R}_k\cdot\mathbf{x}_i=z_k\mathbf{w}_k\cdot\mathbf{x}_i=x_i-h_k$, where $x_i$ is the $x$ coodinate of the $ i $th sample.

%\begin{figure}
%	%\begin{tabular*}{cc}
%	\begin{minipage}{0.48\linewidth}
%		\centerline{\includegraphics[width=3.0cm]{fig3a}}
%	\end{minipage}
%	\hfill
%	\begin{minipage}{.48\linewidth}
%		\centerline{\includegraphics[width=3.0cm]{fig3b}}
%	\end{minipage}
%	\caption{Computing the probability of existing local minima when all weight vectors are parallel.}
%	\label{fig:res}
%\end{figure}

The weights partition the 1D input space into a series of regions $\Omega_j$. Each region lies between two adjacent weight vectors, and in each region $\Omega_j$, $I_{ik} \ (i\in \Omega_j, \ \textup{i.e., sample} \ x_i \ \textup{is located in} \ \Omega_j)$ have constant values, which we denote as $I_{\Omega_jk}$. The total loss can be written as
\begin{equation}\label{eq23}
L=\sum_{\Omega_j} \sum_{i\in \Omega_j} (\sum_{k}{I_{\Omega_jk} x_i - \sum_{k} { I_{\Omega_jk} h_k-y_i})}^2.
\end{equation}

By the global optimality conditions $\frac{\partial L}{\partial h_k}=0 \ (k \in [K])$, we can derive the following linear system,
\begin{equation}\label{eq24}
F\mathbf{h}=\mathbf{f}, 
\end{equation}
where $ \mathbf{h}={(h_1,h_2,\cdots,h_K)}^T $,  $F$ and $\mathbf{f}$ are matrix and vector respectively with the following elements,
\begin{equation*}\label{eq}
F(l,k)=\sum_{\Omega_j} I_{\Omega_jl}I_{\Omega_jk} N_j, \ (l,k\in [K]),
\end{equation*}
\begin{equation}\label{eq25}
\mathbf{f}(l)=\sum_{k} {\sum_{\Omega_j} I_{\Omega_jl}I_{\Omega_jk} \sum_{i\in \Omega_j} x_i-\sum_{\Omega_j} I_{\Omega_jl} \sum_{i\in \Omega_j} y_i}, \ (l\in [K]). 
\end{equation}
$N_j$ is the number of samples in region $\Omega_j$.

Let us discuss the simple case of two weight vectors at first, with normals shown in Fig.1. The analytical solution to the linear system \eqref{eq24} can be easily obtained in this setting. \eqref{eq24} becomes
\begin{equation}\label{eq26}
\left(\begin{matrix}N_1&0\\0&N_3\\\end{matrix}\right)\left(\begin{matrix}h_1\\h_2\\\end{matrix}\right)=\left(\begin{matrix}N_{1+}{\bar{x}}_{1+}+N_{1-}{\bar{x}}_{1-}-N_{1+}+N_{1-}\\N_{3+}{\bar{x}}_{3+}+N_{3-}{\bar{x}}_{3-}-N_{3+}+N_{3-}\\\end{matrix}\right),
\end{equation}
where $N_{1+}$ is the number of positive examples in region $\Omega_1$, ${\bar{x}}_{1+}$ is the average of $x$ coordinates for all positive samples in $\Omega_1$, and so on. Assuming positive and negative classes have equal priors (thus the numbers of positive and negative samples are equal), and denoting the probability of positive (negative) examples lying in region $\Omega_j$ as $P_{j+}$ ($P_{j-}$), the solution to \eqref{eq26} is as follows
\begin{equation}\label{eq27}
\begin{aligned}
x_{\mathbf{w}_{1}^\ast}&=h_1^\ast=\frac{P_{1+}{\bar{x}}_{1+}+P_{1-}{\bar{x}}_{1-}-P_{1+}+P_{1-}}{P_{1+}+P_{1-}}, \\  x_{\mathbf{w}_2^\ast}&=h_2^\ast=\frac{P_{3+}{\bar{x}}_{3+}+P_{3-}{\bar{x}}_{3-}-P_{3+}+P_{3-}}{P_{3+}+P_{3-}},
\end{aligned}
\end{equation}
where $x_{\mathbf{w}_1^\ast}$ is the $ x $ coordinate of $\mathbf{w}_1^\ast$. 

\begin{figure}[ht]
	\vskip 0.2in
	\begin{center}
		\centerline{\includegraphics[width=4.0cm]{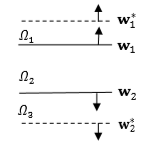}}
		\caption{Two weight vectors with three regions in input space}
		\label{}
	\end{center}
	\vskip -0.2in
	\label{fig1} 
\end{figure}

If there are more than two weight vectors, we need to solve the linear system \eqref{eq24}, with the following $F$ and $\mathbf{f}$ expressed in $\left \{ P_{j+},P_{j-},{\bar{x}}_{j+},{\bar{x}}_{j-} \right \}$ (assuming equal priors for positive and negative classes), 
\begin{equation}\label{E.4}
F\left(l,k\right)=\sum_{\mathrm{\Omega}_j}{I_{\mathrm{\Omega}_{j}l}I_{\mathrm{\Omega}_{j}k}(P_{j+}+P_{j-})},\ \ \ l,k\in[K],
\end{equation}
\begin{equation}\label{E.5}
%\begin{aligned}
\begin{split}
\mathbf{f}\left(l\right)&=\sum_{k}{\sum_{\mathrm{\Omega}_j}{I_{\mathrm{\Omega}_{j}l}I_{\mathrm{\Omega}_{j}k}}(P_{j+}{\bar{x}}_{j+}+P_{j-}{\bar{x}}_{j-}) }\\ &-\sum_{\mathrm{\Omega}_j} I_{\mathrm{\Omega}_{j}l}(P_{j+}-P_{j-}),\ \ \ l\in[K]
\end{split}
%\end{aligned}
\end{equation}

\subsection{Probability of Existing Differentiable Local Minima for 1D Gaussian Data}\label{section3.2}

Assume data samples are drawn from 1D Gaussian distribution. As shown in Fig.1, there exist gaps between $\mathbf{w}^\ast_j$ and $\mathbf{w}_j$. If no samples lie in these gaps, there will be a genuine local minimum in the cells $\left \{ \mathbf{w}_1, \mathbf{w}_2 \right \}$ lie in. Therefore, suppose $N$ samples are i.i.d. drawn, the probability of existing genuine local minima is
\begin{equation}\label{eq28}
P_t={(1-P_g)}^N
\end{equation}
where $P_g$ is the probability of a sample lying in one of the gaps. Denote the gaps as $g_1,g_2,...,g_K$, we use 
$\max _{i} P\left(x \in g_{i}\right)$ to approximate $P_g$ due to
\begin{equation}\label{e8}
P_{g}=P\left(x \in g_{1} \cup g_{2} \cdots \cup g_K\right) \geq \max _{i} P\left(x \in g_{i}\right)
\end{equation}
Since $P_t$ is exponentially vanishing, the probability of existing local minima is very small as long as one of the gaps is large enough such that probability of having samples in it is unnegligible. This conclusion still holds for data samples drawn from other distributions, due to the fact $\mathbf{w}_i^\ast$ and $\mathbf{w}_i$ usually form an intermediate region in which the probability of having samples is nonzero.

The probabilities of existing saddle points and non-differentiable local minim are exponentially vanishing as well due to the gaps. 

We now describe the detailed analytical calculations of $P_g$ and $\left \{ P_{j+},P_{j-},{\bar{x}}_{j+},{\bar{x}}_{j-} \right \}$ for 1D Gaussian distribution. Assume both positive and negative samples are drawn from 1D Gaussian distributions, with means $x_+=1$\ and\ $x_-=-1$ respectively and a standard deviation of 1. Therefore, the probability densities are as follows
\begin{equation}\label{E.6}
P_\pm\left(x\right)=\frac{1}{\sqrt{2\pi}}e^{-\ \frac{{(x-x_\pm)}^2}{2}}
\end{equation} 
Assuming equal priors for positive and negative samples, the probability of a sample lying in gap $g_i$ is
\begin{equation}\label{E.7}
%\begin{aligned} 
\begin{split}
P\left(x\in g_i\right)
&=\frac{1}{2}\left[P\left(x\in g_i\ \right|\ y=1\right)+P\left(x\in g_i\ \right|\ y=-1)] \\
&=\ \frac{1}{2}\left[\int_{g_i}P_+\left(x\right)dx+\int_{g_i}{P_-\left(x\right)dx}\right]\\
&=\frac{1}{2}\left|\int_{x_{\mathbf{w}_i}}^{x_{\mathbf{w}_i^\ast}}{\frac{1}{\sqrt{2\pi}}e^\frac{-\left(x-x_+\right)^2}{2}dx}\right| \\ 
& \ \ \  +\frac{1}{2}\left|\int_{x_{\mathbf{w}_i}}^{x_{\mathbf{w}_i^\ast}}{\frac{1}{\sqrt{2\pi}}e^\frac{-\left(x-x_-\right)^2}{2}dx}\right| \\
&=\frac{1}{2}\left|\Phi\left(x_{\mathbf{w}_{i}^{*}}-x_{+}\right)-\Phi\left(x_{\mathbf{w}_{i}}-x_{+}\right)\right| \\ & \ \ \ +\frac{1}{2}\left|\Phi\left(x_{\mathbf{w}_{i}^{*}}-x_{-}\right)-\Phi\left(x_{\mathbf{w}_{i}}-x_{-}\right)\right|,
%\end{aligned} 
\end{split}
\end{equation}
where $x_{\mathbf{w}_i},x_{\mathbf{w}_i^\ast}$ are the $ x $ coordinates of $\mathbf{w}_i$ and  $\mathbf{w}_i^\ast$ respectively, $\mathrm{\Phi}(x)$ is the cumulative distribution function of standard Gaussian distribution $N(0,1)$. 

When there are multiple weight vectors and consequently multiple gaps, the quantities $\left \{ P_{j+},P_{j-},{\bar{x}}_{j+},{\bar{x}}_{j-} \right \}$ appeared in \eqref{E.4} and \eqref{E.5} for each region $\mathrm{\Omega}_j$ are computed as follows. We will take $ P_{1+} $ and $ \bar{x}_{1+} $ as examples. Using the Gaussian distributions in \eqref{E.6}, we get 
\begin{equation}\label{e9}
\begin{split}
P_{1+}&=\int_{x_{\mathbf{w}_{1}}}^{\infty} \frac{1}{\sqrt{2 \pi}} e^{-\frac{\left(x-x_{+}\right)^{2}}{2}} d x=\int_{x_{\mathbf{w}_{1}}-x_{+}}^{\infty} \frac{1}{\sqrt{2 \pi}} e^{\frac{-x^{2}}{2}} d x \\ 
&= 1-\Phi\left(x_{\mathbf{w}_{1}}-x_{+}\right).
\end{split}
\end{equation}
Using truncated Gaussian distribution, we have
\begin{equation}\label{e10}
\begin{split}
\bar{x}_{1+}
&=\int_{x_{\mathbf{w} 1}}^{\infty} x \cdot \frac{1}{\sqrt{2 \pi}} e^{\frac{-\left(x-x_{+}\right)^{2}}{2}} d x / \int_{x_{\mathbf{w}_{1}}}^{\infty} \frac{1}{\sqrt{2 \pi}} e^{\frac{-\left(x-x_{+}\right)^{2}}{2}} d x\\
&=\frac{\int_{x_{\mathbf{w}_{1}}}^{\infty}\left(x-x_{+}\right) \frac{1}{\sqrt{2 \pi}} e^{\frac{-\left(x-x_{+}\right)^{2}}{2}} d x}{1-\Phi\left(x_{\mathbf{w}_{1}}-x_{+}\right)} \\
& \ \ \ \ +\frac{x_{+} \int_{x_{\mathbf{w}_{1}}}^{\infty} \frac{1}{\sqrt{2 \pi}} e^{\frac{-\left(x-x_{+}\right)^{2}}{2}} d x}{1-\Phi\left(x_{\mathbf{w}_{1}}-x_{+}\right)} \\
&=\frac{1}{\sqrt{2 \pi}} e^{\frac{-\left(x_{\mathbf{w}_{1}}-x_{+}\right)^{2}}{2}} /\left[1-\Phi\left(x_{\mathbf{w}_{1}}-x_{+}\right)\right]+x_{+}
\end{split}
\end{equation}

$P_{1-},{\bar{x}}_{1-}$ can be obtained similarly, and the quantities for regions other than $\mathrm{\Omega}_1$ can be calculated in similar ways, using corresponding intervals for the integrals.

\subsection{Experimental Results for 1D Gaussian Data}\label{section3.3}
In this subsection, we conduct experiments to show how big the probability of existing differentiable local minima is in the whole weight space for 1D Gaussian data. Samples of both positive and negtive classes are drawn from Gaussian distribution with a standard deviation of 1 and means locating at $x_+=1$ and $x_-=-1$ respectively. \textit{N} is set to 100. 

We first consider the case of two weight vectors: $ \mathbf{w}_1 $ and $ \mathbf{w}_2 $. At first, we fix $ x_{\mathbf{w}_2}=0 $ and move $ \mathbf{w}_1 $ in the interval [0, 6]. Fig.2(a)-(c) show three examples of empirical losses w.r.t. $ x_{\mathbf{w}_1} $ obtained by three independant data samplings. In each empirical loss there is clearly a global minimum. We then use \eqref{eq27} to compute $ x_{\mathbf{w}_1^\ast} $ and $ x_{\mathbf{w}_2^\ast} $, and compute the probability of existing local minima with \eqref{eq28}. Fig.2(d) shows how this probability varies with $ x_{\mathbf{w}_1} $. It clearly shows that there is really a high probability of existing local minima at the locations of global minima of empirical losses, and the probability of existing local minima gets sufficiently large when $ x_{\mathbf{w}_1} $ is big enough. These are consistent with the landscapes of empirical losses shown in Fig.2(a)-(c), demonstrating the correctness of our theory on the probability of existing local minima. When $ x_{\mathbf{w}_1} $ is far away from data means, this probability is close to 1 and the loss is almost constant, corresponding to the case of flat plateau local landscape. This can be attributed to the fact that although there is still a gap between $ x_{\mathbf{w}_1} $ and\ $ x_{\mathbf{w}_1^\ast} $, the probability of samples lying in this gap is very low due to the exponentially vanishing nature of Gaussian density when $ x_{\mathbf{w}_1} $ is far away from data means. In other words, almost no samples are activated by $ \mathbf{w}_1 $ in this case, thus moving $ \mathbf{w}_1 $ does not affect the loss. The probability of existing local minima is very low in other places in the weight space due to high probability of having samples in the gap.

We then consider the case in which both weight vectors can move. Fig.2(e) shows the probability of existing local minima when moving  both weights. It is actually the tensor product of the two probabilities obtained by moving $ x_{\mathbf{w}_1} $ and $ x_{\mathbf{w}_2} $ independently. The small peak close to origin corresponds to the global minimum of loss landscape. The probability of existing bad local minima is very low if both weights are not far away from data means. 

Finally, we perform experiments to show the probability of existing differentiable local minima when there are multiple weight vectors. We first test $ K=4 $ weights whose initial locations are [-0.05, 0, 0.05, 0.1] respectively. The normals of the first $ K/2 $ weight vectors are set to $-\mathbf{i}$, and the normals of the last $ K/2 $ weight vectors are set to $\mathbf{i}$. We then shift these weights and solve \eqref{eq24} to get their optimal locations. Due to the hugeness of high-dimensional weight space, we can only visualize some slices of it. Fig.3(a) shows the probability of existing differentiable local minima when moving only the rightmost weight, and Fig.3(b) exhibits the case of moving all weights simultaneously. We then use $ K=60 $ and $ K=200 $ weight vectors respectively that are evenly located in [-3, 3] and shift these weights simultaneously, and solve \eqref{eq24} to get their optimal locations and compute $ P_t $ (the probability of existing genuine local minima) with \eqref{eq28}. These results are given in Fig.3(c) and Fig.3(d) respectively. All these results demonstrate that probability of existing bad local minima is usually very low.

\begin{figure*}[t]
	\centering
	\subfigure[]{
		\label{fig:subfig:onefunction} 
		\includegraphics[width=3.0cm]{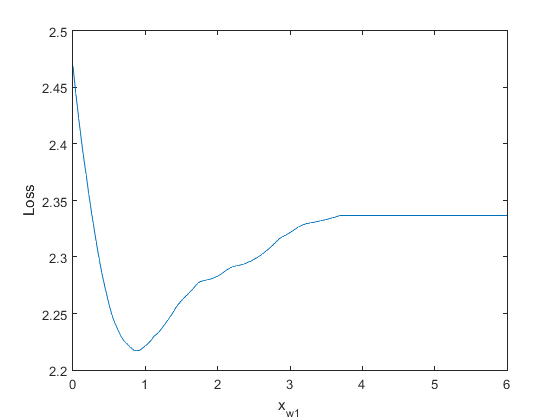}}
	\hspace{0.05cm}
	\subfigure[]{
		\label{fig:subfig:onefunction} 
		\includegraphics[width=3.0cm]{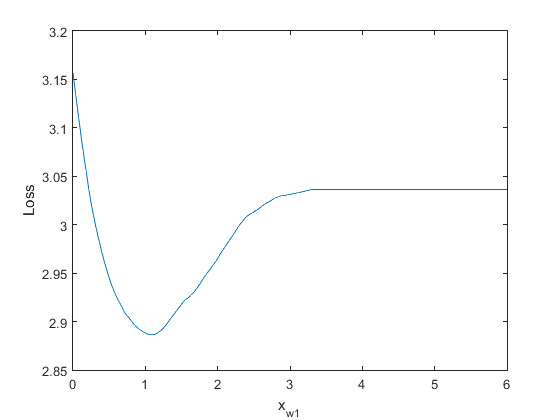}}
	\hspace{0.05cm}
	\subfigure[]{
		\label{fig:subfig:onefunction} 
		\includegraphics[width=3.0cm]{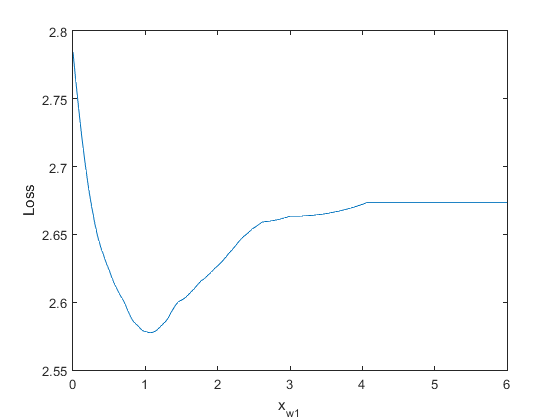}}
	\hspace{0.05cm}
	\subfigure[]{
		\label{fig:subfig:twofunction} 
		\includegraphics[width=3.0cm]{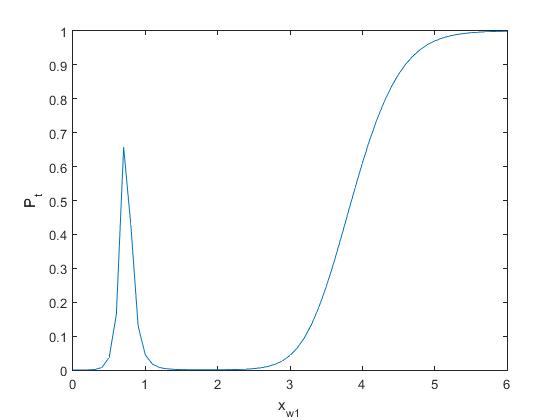}}
	\hspace{0.05cm}
	\subfigure[]{
		\label{fig:subfig:twofunction} 
		\includegraphics[width=3.0cm]{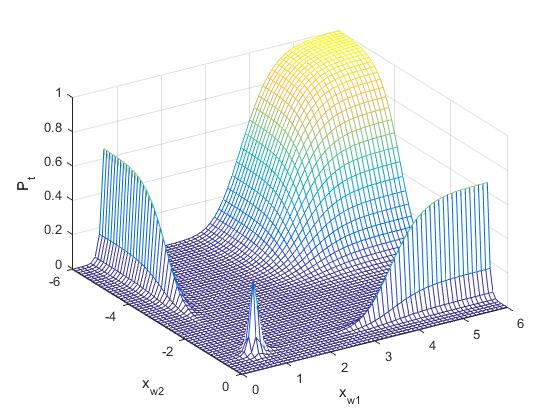}}
	\caption{The probability of existing local minima w.r.t. the locations of two weight vectors. (a)-(c) Three empirical loss landscapes when moving one of the weights. (d) The probability of existing local minima when moving one  of the weights. (e) The probability of existing local minima when moving both weights.}
	\label{fig:twopicture} 
\end{figure*}

\begin{figure*}
	\centering
	\subfigure[4 weights: shifting the rightmost one.]{
		\label{fig:subfig:onefunction} 
		\includegraphics[width=3.5cm]{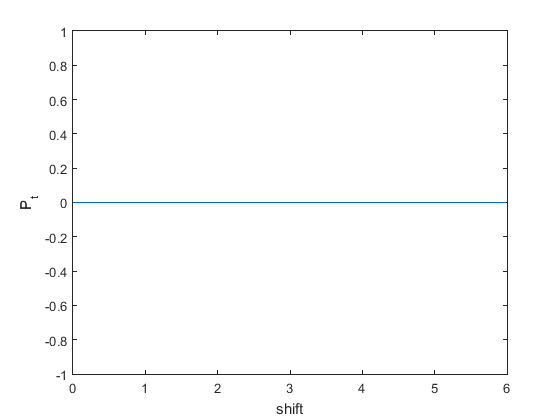}}
	\hspace{0.25cm} % \hspace{0.05cm}
	\subfigure[4 weights: shifting all weights.]{
		\label{fig:subfig:twofunction} 
		\includegraphics[width=3.5cm]{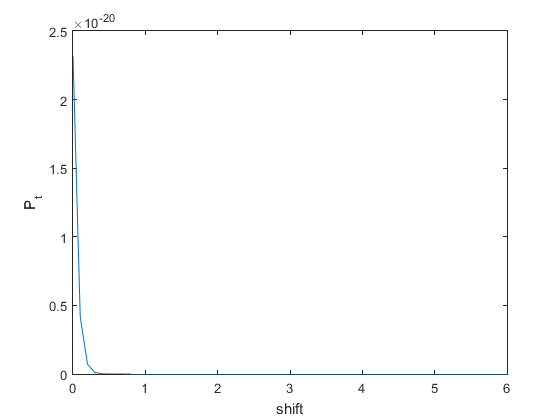}}
	\hspace{0.25cm} % \hspace{0.05cm}
	\subfigure[60 weights: shifting all weights.]{
		\label{fig:subfig:twofunction} 
		\includegraphics[width=3.5cm]{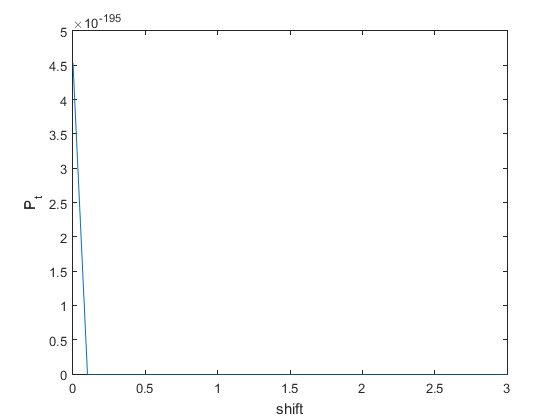}}
	\hspace{0.25cm} % \hspace{0.05cm}
	\subfigure[200 weights: shifting all weights.]{
		\label{fig:subfig:twofunction} 
		\includegraphics[width=3.5cm]{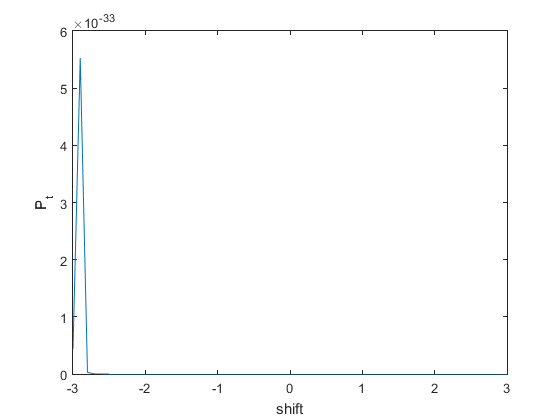}}
	\hspace{0.25cm} % \hspace{0.05cm}
	\caption{The probability of existing local minima w.r.t. the locations of weight vectors}
	\label{fig:twopicture} 
\end{figure*}

\section{Experiments on Higher-dimensional Gaussian data}\label{section4}
\subsection{Dataset and Implementation}\label{section4.1}
In this section, we will present experiments on high-dimensional Gaussian data to probe the existence of bad local minima. Again the purpose is to show how the probability of existing  bad local minima varies in the weight space. However, unlike the 1-dimensional case, it is hard to derive an analytical expression of this probability for higher-dimensional Gaussian data, since finding the regions $\Omega_j$ and computing the integrals of Gaussian distribution over these polyhedron-shaped regions are difficult in high-dimensions. Instead of analytical analysis, we take a sampling approach that works on discrete samples drawn from Gaussian distribution. Both positive and negative samples are drawn from symmetrical multivariate Gaussian distributions. The $ x $ coordinate of positive mean is set to 1 and that of negative mean is set to -1, and all other coordinates of means are  set to 0. Covariance matrices of both distributions are set as identity matrix.

With  Gaussian samples, the optimal weights can be obtained by \eqref{eq7}. We then need to judge whether the optimal weights are genuine using \eqref{eq11} and \eqref{eq12} if they are isolate points. When the optimal weights are in the form of hyperplanes, one has to determine whether the intersection of half-spaces in \eqref{eq13} is null. A naive implementation by computing the cells at first using arrangement of hyperplanes of samples \cite{ComputationalGeometry} and then finding intersections of cells with hyperplanes of optimal weights would be very costly since the arrangment algorithm has a complexity of $ O(N^d) $. In our implementation, we use a much efficient scheme based on linear programming. Since each half-space in \eqref{eq13} is defined by a linear inequality, the problem of judging whether the intersection of half-spaces is null is equivalent to determining whether feasible solutions exist for a linear programming problem with linear inequality constraints in \eqref{eq13} and a dummy linear objective. We use the 'linprog' function in Matlab to perform linear programming with a built-in active-set algorithm. 

The ideal goal is to give the probability of existing  genuine local minima at any point in the weight space. However, since each weight vector is located in a cell, the amount of cells is enormous (see section \ref{section6}) and the number of possible combinations of $ K $ cells weight vectors lie in is exponentially explosive, it is impractical to explore all possible configurations of weight vectors. Therefore, in our experiments we test some typical locations in the weight space. At each location, a random weight matrix composed of $ \left \{\mathbf{w}_j, \ j \in [K] \right \} $ is drawn from Gaussian distribution, which is a common practice when optimizing deep learning models, then the cells weight vectors $ \left \{\mathbf{w}_j, \ j \in [K] \right \} $ lie in are found by computing $ \left \{ I_{ij} \right \} $, and the genuineness of optimal weight vectors is determined by \eqref{eq11} and \eqref{eq12}, or \eqref{eq13}. The location of a weight vector is determined by its bias, so we shift the weight vectors from far distances to near the origin by setting appropriate biases. The initial values of $ \left \{z_j \right \} $ do not matter since $ \left \{ I_{ij} \right \} $ and consequently the defining cells are only determined by $ \left \{\mathbf{w}_j \right \} $.

More specifically, we try some typical locations in the weight space, ranging from locations where all hidden neurons are activated by all samples ($ bias=20 $), locations with small distances to the origin ($ bias=3 $ and $ bias=-3 $) and around the origin ($ bias=0 $), to locations where all hidden neuons are not activated for all samples ($ bias=-20 $). At each weight location, unless otherwise stated, we draw random weight matrix 100 times and report the average results or percentages.  

We also give the probabilities of existing genuine local minima at different locations. After obtaining the optimal solution for a specific weight matrix, we can compute the gaps and consequently the probability of existing  genuine local minima at this location. To circumvent the diffculty of deriving analytical expressions, the probability of a sample lying in a gap is approximated by the ratio of samples in this gap, and the largest probability among all gaps is used to approximate the  probability of existing  genuine local minima, as did in \eqref{e8}. 

Several different combinations of $ K $ (the number of hidden neurons), $ d $ (input dimension) and $ N $ (the number of samples) are tried in our experiments. The results for different combinations are presented in Table \ref{table1}, Table \ref{table2} and Table \ref{table3}. The number of samples is set to $ N=2000 $ for $ K=10 $ and $ K=20 $, and $ N=400 $ for $ K=50 $. In these tables, besides percentage of existing genuine local minima, percentage of them being continuous, and average probability of having samples in gaps ($ P_g $), we also give the average percentage of activated states for hidden neurons at different biases.

Our implementation is in Matlab, and all experiments are conducted on a commodity laptop computer without GPU acceleration.

\begin{table*}[t] %\label{table1}
	\caption{Existence of genuine local minima for Gaussian data ($ d=3, K=10, N=2000 $).}	
	\vskip 0.15in
	\begin{center}
		\begin{small}
			%	\begin{sc}
			\begin{tabular}{ccccc}
				\toprule
				Bias & Genuineness (\%) & Continuousness (\%) & Average $ P_g $ & Activated states (\%) \\
				\midrule
				20  & 0 & 100 & 0.50  & 100  \\
				3   & 0 & 60  & 0.98  & 91.27 \\
				0   & 0 & 3   & 0.92  & 50.66 \\
				-3  & 0 & 86  & 0.96  & 8.04 \\
				-20 & 100 & 100 & 0     & 0  \\				
				\bottomrule
			\end{tabular}
			%	\end{sc}
		\end{small}
	\end{center}
	\vskip -0.1in
	\label{table1}
\end{table*}

\begin{table*}[t]
	\caption{Existence of genuine local minima for Gaussian data ($ d=3, K=20, N=2000 $).}
%	\label{sample-table}
	\vskip 0.15in
	\begin{center}
		\begin{small}
			%	\begin{sc}
			\begin{tabular}{ccccc}
				\toprule
				Bias & Genuineness (\%) & Continuousness (\%) & Average $ P_g $ & Activated states (\%)  \\
				\midrule
				20  & 0 & 100 & 0.51  & 100  \\
				3   & 0 & 91   & 0.99  & 91.67 \\
				0   & 0 & 9   & 0.97  & 50.23 \\
				-3  & 0 & 98   & 0.99  & 7.93 \\
				-20 & 100 & 100 & 0     & 0  \\				
				\bottomrule
			\end{tabular}
			%	\end{sc}
		\end{small}
	\end{center}
	\vskip -0.1in
	\label{table2}
\end{table*}

\begin{table*}[t]
	\caption{Existence of genuine local minima for Gaussian data ($ d=10, K=50, N=400 $).}
%	\label{sample-table}
	\vskip 0.15in
	\begin{center}
		\begin{small}
			%	\begin{sc}
			\begin{tabular}{ccccc}
				\toprule
				Bias & Genuineness (\%) & Continuousness (\%) & Average $ P_g $ & Activated states (\%)  \\
				\midrule
				20  & 0 & 100   & 0.48  & 100  \\
				3   & 0 & 100   & 0.77  & 81.99 \\
				0   & 0 & 100   & 0.73  & 49.33 \\
				-3  & 0 & 100   & 0.77  & 18.38 \\
				-20 & 100 & 100 & 0     & 0  \\				
				\bottomrule
			\end{tabular}
			%	\end{sc}
		\end{small}
	\end{center}
	\vskip -0.1in
	\label{table3}
\end{table*}

\subsection{Results and Analysis}\label{section4.2}

The results in Table \ref{table1}, Table \ref{table2} and Table \ref{table3} show that, when some hidden neurons are activated ($ bias=20, 0, 3, -3 $), there exist no genuine local minima. At these locations, the probability $ P_g $ of having a sample in the gaps between weight vectors and their optimal solutions is big enough such that the probability of existing genuine local minima vanishes. When all hidden neuons are dead for all samples ($ bias=-20 $), the movement of weight vectors does not affect the loss and thus the landscape at this location is a flat plateau with a loss of 1 (since label is either 1 or -1), hence genuine local minima exist. 

In all cases of $ K $, $ d $ and $ N $ at $ bias=20 $, since all hidden neurons are activated, $ I_{ij}=1 \ (i\in [N]; j\in [K])$ and thus $ A $ in \eqref{eq6} is rank-deficient, the local minima will be continous according to subsection \ref{section2.1}. At $ bias=-20 $, the landscape is a flat plateau and thus is continous as well. These statements are verified in Table \ref{table1}, Table \ref{table2} and Table \ref{table3}. 

Comparing Table \ref{table1} and Table \ref{table2}, the percentages of activated neurons remain roughly the same due to same data and same biases. Moreover, for the cases of $ bias =3, 0, -3 $, more local minima are continuous when $ K $ is bigger. The reason is as follows. In these two tables there is $ N>Kd $, hence according to subsection \ref{section2.1}, the local minima will be continuous if $ rank (A) < Kd $ and isolated if $ rank (A) = Kd $. With the increase of $ K$, more columns in matrix $ \left ( I_{ij} \right )\in \mathbb{R}^{N\times K} $ will equal 1s by the almost constant percentage of activated states. As a result, it is more likely to have $ rank (A) < Kd $ because of identical columns in $ A $, leading to more continuous local minima.

In Table \ref{table3}, all local minima are continous due to $ N<Kd $. This case corresponds to over-parameterization and the locations of some weight vectors can move freely without changing the loss.

\section{Experiments on Real Datasets}\label{section5}
\subsection{Datasets and Implementation}\label{section5.1}

In order to show the theoretical predictions on the existence of genuine local minima and verify their correctness for real datasets, we perform experiments on two datasets commonly used in machine learning community: MNIST and CIFAR-10 images. The MNIST dataset consists of $ 28\times 28 $ images of handwritten digits, and the CIFAR-10 dataset comprises $ 32\times 32 $  color images of objects belonging to 10 classes including airplane, automobile and bird etc. We extract the first 100 samples of digit 0 and digit 1 respectively from MNIST training data, and thus construct a binary classification problem with 200 samples. For CIFAR-10 dataset, a binary classification problem with 200 samples is formed similarly with images from the airplane and automobile classes. We convert the CIFAR-10 color images into gray images to reduce the input dimension and consequently computational overhead. Image pixel values are scaled from [0, 255] to [0, 1]. The number of hidden neurons is set to $ K=10 $ for both datasets. 

In order to determine whether a local minimum is genuine when it is a hyperplane, we again use the linear programming approach to judge whether the intersection of half-spaces is null. For MNIST data with $ N=200 $ samples and $ K=10 $ hidden neurons, each trial of the Matlab 'linprog' function takes about 1.7 miniutes to converge on our commodity laptop computer without GPU acceleration, and at each bias we run 100 trials each with different random weight vectors. For CIFAR-10 data, it takes about 1.5 hours to converge with $ N=200 $ and $ K=10 $, and we run 10 trials at each bias to save time. At each bias, the percentage of existing genuine local minima, percentage of them being continuous, and the average percentage of activated states for hidden neurons are reported.

Besides theoretical prediction on the existence of genuine local minima, we also design and implement a method to verify whether genuine local minima really exist and compare the results with our theoretical predictions. We use gradient descent to search for local minima. Starting from random $ \left \{z_j \right \} $ and $ \left \{\mathbf{w}_j \right \} $ with each weight vector $ \mathbf{w}_j $ lying in a certain cell, we run gradient descent to update $ \left \{\mathbf{w}_j \right \} $ and $ \left \{z_j \right \} $. The gradients are given as follows for squared loss function: $\forall  j \in[K]$,  
\begin{equation}\label{A.3}
\frac{\partial L}{\partial z_j}=\frac{\partial L}{\partial\mathbf{R}_j}\cdot{\mathbf{w}_j}, \ \ \ \frac{\partial L}{\partial\mathbf{w}_j}=\frac{\partial L}{\partial\mathbf{R}_j}\cdot {z_j},
\end{equation}
\begin{equation}\label{C.5}
\begin{split}
\frac{\partial L}{\partial\mathbf{R}_j}=\frac{2}{N}\sum_{i=1}^{N}{\left(\sum_{k=1}^{K}{I_{ik}\mathbf{R}_k\cdot\mathbf{x}_i}-y_i\right)\cdot I_{ij}\mathbf{x}_i}     
\end{split}	
\end{equation}	

Once gradient descent moves across cell boundaries, that is, $\exists i\in [N], \exists j\in [K]$, $\left ( \mathbf{w}_j \cdot \mathbf{x}_i \right ) \left ( \mathbf{{w}'}_j \cdot \mathbf{x}_i \right ) \leq 0 $, where $ \mathbf{{w}'}_j $ is the updated weight vector, we can conclude that there are no genuine local minima in the cells specified by $ \left \{\mathbf{w}_j \right \} $. Otherwise, if the updated weight vectors are still inside their defining cells and the gradient magnitudes are sufficiently small, we then conclude that there exist genuine local minima. Therefore, the results of gradient descent can be directly compared with theoretical predictions and verify their correctness. In our implementation, the stepsize of gradient descent is set to $ 10^{-6} $, and the threshold of gradient magnitude for identifying existence of local minima is set to  $ 10^{-3} $.

\begin{table*}[t]
	\caption{Existence of genuine local minima for MNIST data ($ d=784, K=10, N=200 $).}
%	\label{sample-table}
	\vskip 0.15in
	\begin{center}
		\begin{small}
			%	\begin{sc}
			\begin{tabular}{ccccc}
				\toprule
				Bias & Genuineness (\%) & Continuousness (\%) & Trapped in cells (\%) & Activated states (\%) \\
				\midrule
				20  & 0 & 100   & 0 & 98.70  \\
				3   & 0 & 100   & 0 & 62.96 \\
				0   & 0 & 100   & 0 & 52.19 \\
				-3  & 0 & 100   & 0 & 37.00 \\
				-50 & 100 & 100 & 100 & 0  \\				
				\bottomrule
			\end{tabular}
			%	\end{sc}
		\end{small}
	\end{center}
	\vskip -0.1in
	\label{table4}
\end{table*}

\begin{table*}[t]
	\caption{Existence of genuine local minima for CIFAR-10 data ($ d=1024, K=10, N=200 $).}
%	\label{sample-table}
	\vskip 0.15in
	\begin{center}
		\begin{small}
			%	\begin{sc}
			\begin{tabular}{ccccc}
				\toprule
				Bias & Genuineness (\%) & Continuousness (\%) & Trapped in cells (\%) & Activated states (\%)  \\
				\midrule
				20  & 0 & 100   & 0 & 86.38  \\
				3   & 0 & 100   & 0 & 57.09 \\
				0   & 0 & 100   & 0 & 49.94 \\
				-3  & 0 & 100   & 0 & 42.47 \\
				-80 & 100 & 100 & 100 & 0  \\				
				\bottomrule
			\end{tabular}
			%	\end{sc}
		\end{small}
	\end{center}
	\vskip -0.1in
	\label{table5}
\end{table*}

\subsection{Results and Analysis}\label{section5.2}
The results for MNIST and CIFAR-10 datasets are given in Table \ref{table4} and Table \ref{table5} respectively. They show that, except the case where all hidden neuons are dead for all samples ($ bias=-50 $ and $ bias=-80 $ respectively), there exist no genuine local minima. This is again due to the fact that there usually have samples in the gaps between weight vectors and their optimal solutions. When all hidden neuons are dead for all samples, the landscape is a flat plateau with a loss of 1 and genuine local minima exist. There must be someplaces where genuine local minima corresponding to global minima exist. However, the possibility of hitting these locations by randomly sampling weight matrix is very low, thus they are missed by the typical locations in Table \ref{table4} and Table \ref{table5}.   

In Table \ref{table4} and Table \ref{table5}, we have $ N<Kd $, hence the local minima (whether being genuine or not) are all continous according to subsection \ref{section2.1}, corresponding to the over-parameterization case.

The columns 'trapped in cells' in Table \ref{table4} and Table \ref{table5} give the percentages of gradient descent trapped in their starting cells at different biases. They are all zeros for those cases no genuine local minima exist by theoretical predictions, indicating the consistency between our theoretical predictions and experimental results on the existence of genuine local minima. For the case all hidden neurons are dead for all samples ($ bias=-50 $ for MNIST, $ bias=-80 $ for CIFAR-10), the gradient magnitudes are zero and gradient descent is trapped, again coinciding with the theoretical predictions of flat plateau landscape . 

We did not try much larger number of hidden neurons and samples for real datasets due to limitation of computation resources. However, the conclusion of no bad local minima almost everywhere does not change since usually there are samples in the gaps between weights and their optimal solutions. Since the goal of this work is to understand the existence of local minima, we also did not put much effort into performance improvement.

\section{Count and Size of Differentiable Cells}\label{section6}
For one-hidden-layer ReLU networks, samples $ (\mathbf{x}_i,\ i\in [N]) $ partition the weight space into a number of convex cells. After understanding the existence of local minima inside cells, we now turn to explore the following questions to give a more complete picture of local loss landscapes: how many cells are there and how big are they? As shown in our previous work \cite{globallosslandscape_part1}, the shape of each cell is a polyhedron since it is formed by intersection of hyperplanes of samples. \cite{arrangements} shows that in $ d $-dimensional input space, an arrangement of $ N $ hyperplanes of samples in general can generate $\sum_{i=0}^{l}\left(\begin{array}{l}N \\ i\end{array}\right)$ cells. Thus, even for a small dataset of 1,000 training examples in 10-dimensional input space, the number of cells in the whole weight space will be as high as $ 1.9 \times 10^{13} $, which makes it impractical to search all the cells exhaustively.

The size of cells tells us how far a local search can go before changing the activation pattern of ReLU neurons. We carry out experiments to explore the average diameter of cells. By diameter, we mean the largest possible Euclidean distance between two points in a cell. We take a sampling approach to calculate the diameters. Starting from a weight vector $ \mathbf{w} $ located in a specific cell, we draw a number of random directions and go along these directions in two opposite ways until hitting the hyperplanes of samples confining the cell. Specifically, for a direction $ \mathbf{d}_j $ and sample $ \mathbf{x}_i $, we can compute the distance from $ \mathbf{w} $ to $ \mathbf{x}_i $ along $ \mathbf{d}_j $. When the ray of  $ \mathbf{d}_j $  intersects with the hyperplane of $ \mathbf{x}_i $, there is $\left(\mathbf{w}+s_{i j} \mathbf{d}_{j}\right) \cdot \mathbf{x}_i=0$, where $ s_{i j} $ is the distance from $ \mathbf{w} $ to the hyperplane of $ \mathbf{x}_i $ along $ \mathbf{d}_j $. Therefore $s_{i j}=-\frac{\mathbf{w} \cdot \mathbf{x}_{i}}{\mathbf{d}_{j} \cdot \mathbf{x}_{i}}$. The distance from $ \mathbf{w} $ to the nearest sample along direction $ \mathbf{d}_j $ is $s1_{j}=\min _{i}\left\{s_{ij} | s_{ij} \geqslant 0\right\}$, and that along direction $-\mathbf{d}_j $ is $s2_{j}=\left |\max _{i}\left\{s_{ij} | s_{ij} < 0\right\}  \right |$. Finally, by finding the largest total distance among all directions $\max _{j}\left(s1_{j}+s2_{j}\right)$, we get the diameter of the cell in which $ \mathbf{w} $ resides. 

The random directions are drawn from uniform distribuiton and the number of directions is set to be ten times of input dimension. We draw uniformly 1000 random weight vectors and calculate the diameters of the cells in which they lie. The open cells are excluded since their diameters tend to be infinite, and the average diameter of remaining cells is reported. We run experiments on several datasets including MNIST, CIFAR-10, and Gaussian data with different input dimensions. For each dataset, we try with two data sizes: $ N=2000 $ and $ N=200 $. The results are reported in Table \ref{table6}.  

\begin{table*}[t]
	\caption{Diameters of differentiable regions for various datasets.}
%	\label{sample-table}
	\vskip 0.15in
	\begin{center}
		\begin{small}
			%	\begin{sc}
			\begin{tabular}{lcccr}
				\toprule
				Dataset & Diameter ($ N $=2000) & Diameter ($ N $=200) \\
				\midrule
				MNIST    & 0.366& 2.687 \\
				CIFAR-10 & 3.374& 6.443\\
				Gaussian (d=3)    & 0.009& 0.094 \\
				Gaussian (d=10)    & 0.009& 0.093 \\
				Gaussian (d=100)     & 0.034& 0.351\\
				Gaussian (d=300)   &0.068&0.678\\	   				
				\bottomrule
			\end{tabular}
			%	\end{sc}
		\end{small}
	\end{center}
	\vskip -0.1in
	\label{table6}
\end{table*}

Table \ref{table6} shows that the size of cells is unsually small compared with that of data bounding box (each dimension is in [0, 1] for MNIST and CIFAR-10 datasets), thus when moving through the weight space, activation pattern of ReLU neurons changes frequently. Table \ref{table6} also shows that the size of cells is getting bigger when the number of samples getting smaller. This is because the space between samples becomes bigger for sparser dataset. Moreover, one can see that for Gaussian data the size of cells becomes bigger for higher-dimesional inputs. This is attributed to the fact that Euclidean distance becomes bigger with dimension.  

\section{Related Work}\label{section7}

There have been some experimental studies on the loss landscape of DNNs and its visualization \cite{Goodfellow,Poggio,EssentiallyNoBarriers,modeconnectivity,Visualizing,largscalelandscape}. \cite{Goodfellow} showed that the linear path from initialization to the optimized solution has a monotonically decreasing loss along it. In this work, for one-hidden-layer ReLU networks, we explained theoretically and verified experimentally that there are no bad differentiable local minima almost everywhere, verifying the conclusion of \cite{Goodfellow}.

\cite{EssentiallyNoBarriers,modeconnectivity} revealed that global minima usually constitute continuous regions for over-parameterized networks. In this work, we give a theoretical explaination of continuous minima for one-hidden-layer ReLU networks, and point out that continuous global minima are actually in the form of hyperplanes. 

The experimental study of \cite{SafranShamir18} showed that spurious local minima are common for two-layer networks where the label is generated by a teacher network with unknown parameters. Their objective function is different than ours and thus the results can not be compared directly. It would be interesting to explore whether our theory of local minima can be applied to the student-teacher two-layer networks. 

Theoretically, \cite{SafranShamir16} computed the probability of getting trapped in the cells that contain global minima during initialization, whereas our theory considered how the probability of existing genuine local minima varies in the whole weight space.

See part 1 of this work \cite{globallosslandscape_part1} for more works related to loss landscape. 

\section{Conclusions and Future Work}

In this paper, based on our theory of local minima for one-hidden-layer ReLU networks in \cite{globallosslandscape_part1}, we first analyzed how the probability of existing genuine differentiable local minima varies in the whole weight space for 1D Gaussian data. We showed that this probability is very low in most regions. We then implemented our theory of existing genuine local minima with linear programming and used it to predict whether bad local minima exist for higher-dimensional Gaussian data, MNIST and CIFAR-10 datasets, and find that there are no bad local minima almost everywhere in weight space once some hidden ReLU neurons are activated. These theoretical predictions were verified by showing experimentally that gradient descent was not trapped in the starting cells at different locations. These theoretical and experimental results explain, from the perspective of loss landscape, why local search based methods can optimize one-hidden-layer ReLU networks of any size and any input dimension if initialized at appropriate locations.  

In future work, we are intertested in conducting experiments on real datasets to explore the existence of non-differentiable local minima and saddle points. We also plan to study the global loss landscape of deep ReLU networks.

% or


%\appendices
%% you can choose not to have a title for an appendix
%% if you want by leaving the argument blank
%\section{}
%% use section* for acknowledgment
%\section*{Acknowledgment}

\ifCLASSOPTIONcaptionsoff
  \newpage
\fi



%\bibliographystyle{IEEEtran}

\bibliographystyle{plain}
\bibliography{Understanding_global_landscape_ReLU}

%
%\begin{thebibliography}{1}
%\end{thebibliography}


%\end{IEEEbiography}
%% insert where needed to balance the two columns on the last page with
%%\newpage


%\vfill


\end{document}